%% file: physxnet_camready.tex
\documentclass[10pt,twocolumn,letterpaper]{article}

\usepackage{3dv}
\usepackage{times}
\usepackage{epsfig}
\usepackage{graphicx}
\usepackage{amsmath}
\usepackage{amssymb}

\usepackage{caption}
\usepackage{makecell}
\usepackage[caption = false]{subfig}
\usepackage{pifont}
\usepackage{booktabs}

\input{newcommands2.tex}

\usepackage[pagebackref=true,breaklinks=true,colorlinks,bookmarks=false]{hyperref}

\threedvfinalcopy 

\def\threedvPaperID{0313} 

\ifthreedvfinal\fi
\begin{document}

\title{PhysXNet: A Customizable Approach for Learning \\ Cloth Dynamics on Dressed People}


\author{Jordi Sanchez-Riera, Albert Pumarola and Francesc Moreno-Noguer \\
Institut de Rob\`{o}tica i Inform\`{a}tica Industrial, CSIC-UPC\\
08028, Barcelona, Spain\\
{\tt\small \{jsanchez,apumarola,fmoreno\}@iri.upc.edu}
}


\twocolumn[{ 

  \maketitle
  \thispagestyle{empty}
  
  \begin{center}
  \vspace{-3mm}
    \includegraphics[width=0.98\linewidth]{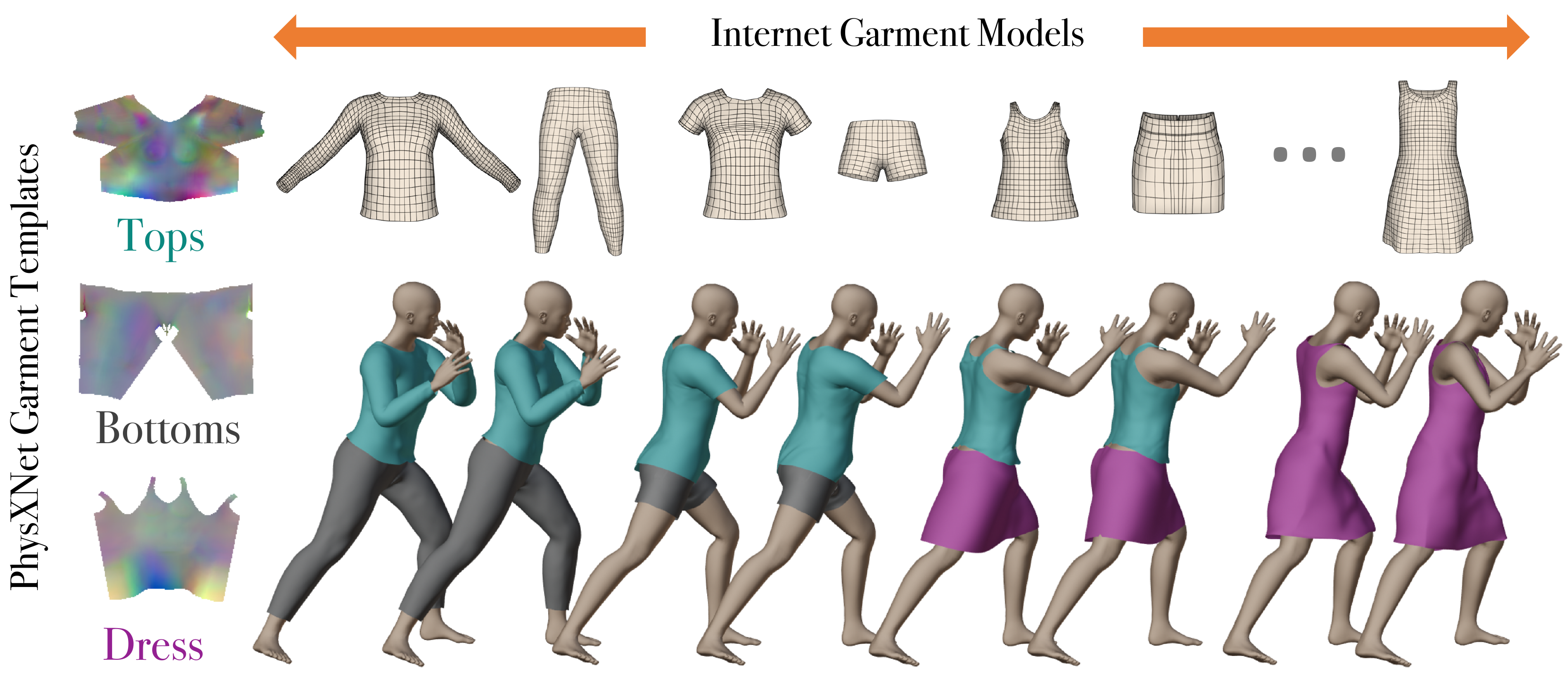} \\
    \captionof{figure}{\small{{\bf PhysXNet} enables predicting cloth dynamics from skeletal motion of the person. We parameterize clothes using generic UV maps that can encapsulate different types of garments. The UV maps on the left of the figure have the capacity to generate all the variety of clothes shown in the top. Specific garments are picked and their dynamic is predicted using PhysXNet and used to dress the moving character.}}\label{fig:teaser}
  \end{center}

}]

\begin{abstract}
\vspace{-3mm}
We introduce PhysXNet, a learning-based approach to predict the dynamics of deformable clothes given 3D skeleton motion sequences of humans wearing these clothes. The proposed model is adaptable to a large variety of  garments and changing topologies, without need of being retrained.
Such simulations are typically carried out by physics engines that require manual human expertise and are subject to computationally intensive computations. PhysXNet, by contrast, is a fully differentiable deep network that at inference is able to estimate the geometry of dense cloth meshes in a matter of milliseconds, and thus, can be readily deployed  as a layer of a larger deep learning architecture.
This efficiency is achieved thanks to the specific parameterization of the clothes we consider, based on 3D UV maps encoding  spatial garment displacements.  The problem is then formulated as a mapping between the human kinematics space (represented also by 3D UV maps of the undressed body mesh) into the clothes displacement UV maps, which we learn using a conditional GAN with a discriminator that enforces feasible deformations. We train simultaneously our model for three garment templates, tops, bottoms and dresses for which we simulate deformations under $50$ different human actions. Nevertheless, the UV map representation we consider allows encapsulating many different cloth topologies, and at test we can simulate garments even if we did not specifically train for them. A thorough evaluation demonstrates that PhysXNet delivers cloth deformations very close to those computed with the physical engine, opening the door to be effectively integrated within deep learning pipelines. 
\end{abstract}

\section{Introduction}
High-fidelity animation of clothed humans is the key for a wide range of applications in \eg AR/VR, 3D content production and virtual try-on. One of the main challenges when generating these animations is to create realistic cloth deformations with plausible wrinkles, creases, pleats, and folds. Such simulations are typically carried out by physics engines that model clothes via meshes with neighboring vertices connected using spring-mass systems. Unfortunately, these simulators need to be fine-tuned by a human expert and are subject to computationally intensive processes to calculate collisions between vertices. These two limitations prevent their deployment as a layer of a larger deep learning architecture.

With the advent of deep learning there have been a number of learning-based approaches that attempt to emulate the physical engines using differentiable networks~\cite{laehner2018deepwrinkles, Gundogdu-ICCV-2019, vidaurre2020virtualtryon, Patel_2020_CVPR}. These approaches, however, propose models that are focused on specific clothes, and need to be retrained to simulate novel garments.
Most recent approaches \cite{Patel_2020_CVPR, vidaurre2020virtualtryon, li2020deep} build upon
an MLP architecture conditioned on body pose, shape and garment style. While the style parameter allows certain control on cloth attributes (\eg  sleeve length, size and fit), the range of variation is still fairly reduced. 

In this paper, we present PhysXNet, a method to predict cloth dynamics of dressed people that is adaptable to different clothing types, styles and topologies without need of being retrained. For this purpose we build upon a simple but powerful representation based UV maps encoding cloth displacements. These UV maps are carefully designed in order to simultaneously encapsulate many different cloth types (upper body, lower body and dresses) and cloth styles (\eg from long-sleeve to sleeve-less T-shirts). Given this representation, we then formulate the problem as a mapping between the human body kinematic space and the cloth deformation space. The input human kinematics are similarly represented as UV maps, in this case encoding body velocities and accelerations. Therefore, the problem boils down to learning a mapping between two different UV maps, from the human to the clothing, which we do using a conditional GAN network. 

In order to train our system we build a synthetic dataset with the Blender physical engine, consisting of 50  skeletal  actions and a human wearing three different garment templates: tops, bottoms and dresses. The results show that PhysXNet is then able to predict very accurate cloth deformations for clothes seen at train, while being also adaptable to clothes with a other topologies with a simple UV mapping.  

Our key contributions can be summarized as follows:
\begin{itemize}
\setlength{\itemsep}{0pt}
\item A model that is able to predict simultaneously deformations on three garment templates. 
\item A garment template representation by means of UV maps that allows us to easily map different cloth topologies onto the these templates.
\item A differentiable network that can be integrated into larger deep learning pipelines.
\item A new dataset of physically plausible cloth deformations with 50 human actions and 3 garment templates: tops, bottoms, and dresses. 
\end{itemize}

\section{Related Work}

Estimating the deformation of a piece of cloth when a human is striking a pose or performing an action is a very difficult task. While estimating cloth deformation has been traditionally addressed by model-based approaches~\cite{Moreno_cvpr2009,Sanchez_cvpr2010,Moreno_pami2013}, recent deep learning techniques build upon data-driven methods. These datasets, however, usually ignore cloth deformation physics, producing unrealistic renders. This problem is generally addressed by obtaining the data from registered scans or including cloth simulation engines into the data generation process. Below, we briefly review different methods and datasets used to achieve realistic cloth and body reconstructions.



\noindent{\bf Synthetic Datasets.} One of the main problems when generating a dataset is to obtain natural cloth deformations when a human is performing an action. Scan based approaches \cite{SimulCap19, habermann20deepcap, SIZER_Dataset} have the advantage that can capture every cloth detail without having to worry about cloth physical models, however, the main drawback is that they need of dedicated hardware and software to process all the data. On the other side, synthetic based approaches \cite{varol17_surreal, pumarola20193dpeople, bertiche2020cloth3d} can be easily annotated and modified, but have the trouble of obtaining realistic cloth deformations. Cloth behaviour has to be tuned for every different 3D model and for each action requiring some professional expertise. Recent cloth physical engines can achieve very natural cloth behaviors \cite{Koh_2015_VDA, Tang2016Cama, Tang2018TOG}, even for complex meshes, which makes the synthetic simulation a good competitor for the scanned data. In our work, we generate a synthetic dataset with a parametric human 3D model \cite{Makehuman} and use Blender \cite{Blender} cloth engine as a cloth simulator. We create high quality cloth deformations for three garment templates over $50$ motion sequences.


\noindent{\bf Data driven cloth deformations.} Using the generated datasets either from scans or synthetic data, a big part of the research concentrate in achieve high detailed cloth deformations with tailored designed networks \cite{ma20autoenclother, Yang2018garments, Gundogdu-ICCV-2019, bhatnagar2019mgn, laehner2018deepwrinkles, zhang2020deep}, GANs \cite{shen2020garmentgeneration, Kato2019clothes} or even more recently with implicit functions \cite{corona2021smplicit}. These methods assume each cloth deformation frame is independent from each other and just concentrate to obtain reliable reconstructions in still images.
Other methods go one step further and they try to infer the cloth deformation given a human pose and shape  \cite{jin2018pixelbased, Geng2020tshirt, pumarola2020d} obtaining very convincing results. In some other cases \cite{vidaurre2020virtualtryon, WangEtAl-SiggraphAsia-2019, santesteban2021garmentcollisions}, the cloth size and style also can be adjusted changing some statistical model weights, which allows more flexible simulations. However, these methods are designed to deal with a single cloth at a time, and cloth dynamics generated by body movements are ignored.

\noindent{\bf Physics based cloth deformations.} Above methods reason about cloth geometry to obtain plausible cloth deformations, but ignore the underlying physics of the cloth, which can help to achieve more natural deformations. This is especially true when the cloth deformations are affected by the motion of the body. Using the physics information obtained from a dataset, different networks \cite{Guan-DRAPE-2012, garmentdesign_Wang_SA18, Santesteban-EG-2019} are able to simulate cloth wrinkles and deformations given a body pose and shape. Tailornet \cite{Patel_2020_CVPR}, extends this work and allows for a cloth style variations obtained from a base template model. While these methods are designed to be optimal on a T-shirt cloth, other cloth garments can be also estimated \cite{li2020deep, santesteban2021garmentcollisions}.
All simulations are achieved using a dedicated network per cloth garment, which makes these methods not very flexible in case our cloth mesh is different that the one they used for train. Moreover, a human model usually wears more than a single cloth garment, which means that these methods need to use different networks for the different garments and make them more difficult to integrate in a more larger pipeline.

\begin{figure*}[t]
\begin{center}
   \includegraphics[width=0.99\linewidth]{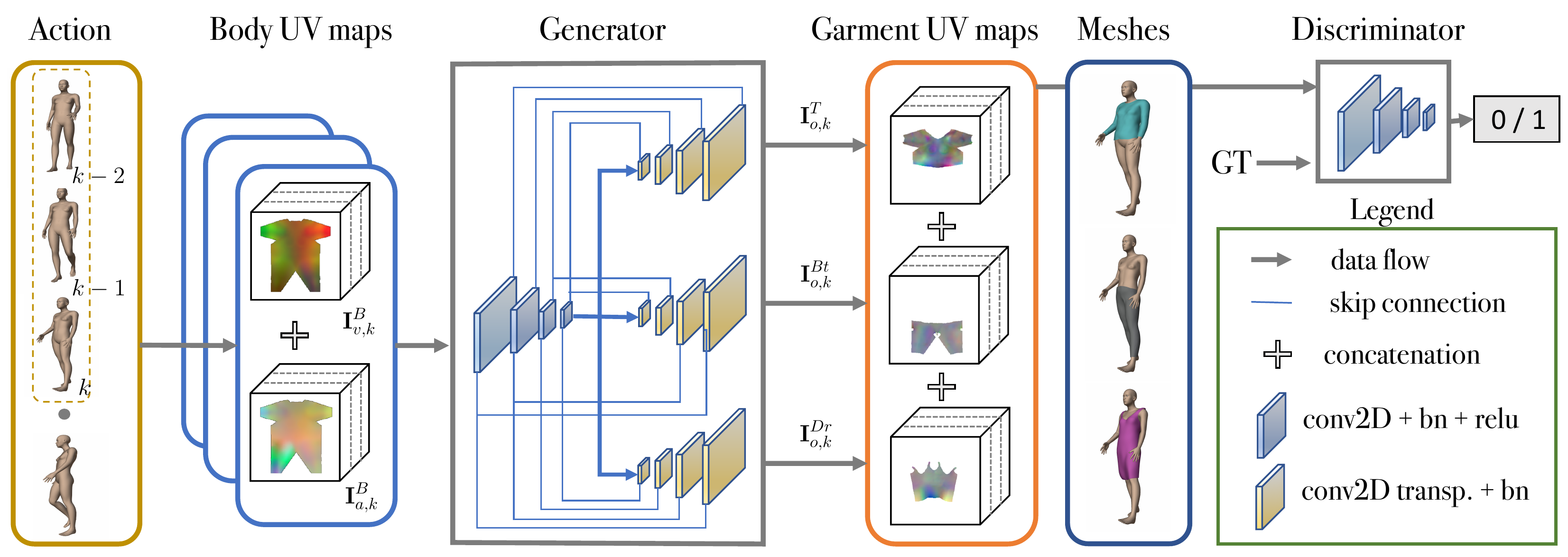}
\end{center}
\vspace{-0.5cm}
   \caption{\textbf{PhysXNet pipeline}. Given a sequence of human body motions without wearing any cloth, the velocities and accelerations of the body surface are calculated and registered in UV maps. The PhysXNet network receive the current body UV maps with the two previous body UV maps to generate three different garment estimates that encodes the offset of the garment respect to the body. These estimates are learnt with an adversarial training. Finally, a given cloth garment is projected to the estimated UV maps to obtain the final cloth deformation. The discriminator of the network is only used in the training stage.}
\label{fig:pipeline}
\vspace{-3mm}
\end{figure*}

\section{PhysXNet}

\subsection{Problem Formulation}
Physics-based engines model clothes using spring-mass models. In an oversimplification of how a simulation is performed, we can understand that the force (and hence the displacement) that receives each of these spring-mass elements is estimated as a function of the the skeleton velocities and accelerations. Building upon this intuitive idea we formulate the problem of predicting cloth dynamics as a regression from current and past body velocities and accelerations to cloth-to-body offset displacements. We encapsulate all these information by means of UV maps.

Concretely, assume we have a sequence of training frames, consisting of 3D body meshes and their corresponding cloth meshes. For a specific frame $k$, we know the body and cloth UV maps that transform 3D points on the surface to 2D points on a planar domain. Let $\bI^B_{v,k}$ and $\bI^B_{a,k}$ denote the body ($B$) UV maps of size $W\times H\times 3$ encoding the velocity (v) and accelerations (a) of the surface points. Similarly, let $\bI^C_{o,k}$ be the cloth ($C$) UV map, also with size $H\times W \times 3$, encoding the offset (o) of the cloth surface points w.r.t. the body surface.  We consider three different cloth templates, tops ($T$), bottoms ($Bt$) and dresses ($Dr$), that is, $C=\{T,Bt,Dr\}$.

Given this notation, we can formulate our problem as that of learning the mapping $\mM : \bX \to \bY$, where $\bX = \{\bI^B_{v,k-2:k},\bI^B_{a,k-2:k}\}$ are the velocities and accelerations of the body surface points in the frames $k-2$, $k-1$ and $k$; and $\bY=\bI^C_{o,k}$ are the garment offsets at the current frame $k$.

\subsection{Model}
Fig.~\ref{fig:pipeline} shows an schematic of the PhysXNet pipeline. Given a sequence of human body motions, the UV maps for body velocities and accelerations are computed in triplets and passed to the network in order to infer the UV maps of the cloth offsets for the current evaluated frame. Then, the vertices of a given garment are projected to the correspondent UV garment map to obtain the offsets respect to the body surface point for each one of the vertices and hence, the final position for the garment cloth.

The PhysXNet network is trained with two separate models where, a generator model produce samples of the UV garment maps, and a disciminator model tries to determine whether these samples are real or fake. Then, it starts a Minimax strategy game \cite{Goodfellow-NIPS-2014} with the generator trying to "fool" the discriminator, and the discriminator trying to "catch" the generator wrong samples.

Thus, the discriminator is trained in a supervised manner, where the input data from the generator should return $D(G(\mathbf{X})) = 0$ and the input real data should return $D(\mathbf{Y}) = 1$. The loss $\mathcal{L}_{adv}$ of the discriminator is given by a sigmoid cross entropy for the real and generated classes:
\begin{align}
    \mathcal{L}_{adv} &= \mathbb{E}_y [log(D(\mathbf{X}))] + \mathbb{E}_x[log(1-D(G(\mathbf{X})))] \label{eq:disc_loss} 
\end{align}

The generator is trained to produce data output as similar as possible to the ground truth data $\mathbf{Y}$. In the generator loss $\mathcal{L}_G$ is used a regularization term, that ensures that generated garment UV maps $\hat{\mathbf{I}}^C_{o,k}$ stay close to the ground truth garment UV maps $\mathbf{I}^C_{o,k}$:
\begin{align}
    \mathcal{L}_G &= \mathbb{E}_x [1 - log(D(G(\mathbf{X})))] + \lambda_{L_1} * | \mathbf{I}^C_{o,k} - \hat{\mathbf{I}}^C_{o,k} |_1 \label{eq:gen_loss} 
\end{align}
where $\lambda_{L_1}$ is a parameter that controls the weight of the regularization term. Note that we use $L1$ metric as we find that produces slightly better results and a more stable training than a $L2$ metric.

\noindent{\bf Architecture.} The generator $G(\mathbf{X})$ is designed as an encoder-decoder network. The encoder network receives the body velocities and accelerations UV maps for the current and previous two frames of the motion sequence. Then the "body" encoder is connected to a "garment" decoder, one for each garment template, that returns the offsets positions of the garment respect to the body. As the garment offsets have different behaviour depending of the template, is necessary to have a different decoder for each garment template. 

Each encoder layer is composed by a 2D convolution that sub-sample the input into a half size, a batch normalization and a ReLU function. Each decoder layer is composed by a transposed 2D convolution that doubles the size of the input, and a batch normalization layer. We use four encoder and decoder layers with skip connections as in the UNet network \cite{Ronnegerger2015unet}. 
The discriminator $D(G(\mathbf{X}))$ is a PatchGAN decoder with a binary output taken from \cite{pix2pix2017}. The use of a discriminator helps the network to produce more smooth UV maps with no abrupt changes between close pixels.

\section{Generate training data}

\subsection{Dataset}
\label{sec:dataset}
The dataset is generated using Avatar Blender add-on \cite{sanchezriera2021avatar}. 
This add-on is based on Makehuman \cite{Makehuman} open source 3D human model and it is completely integrated to the rendering software Blender \cite{Blender}. It includes a parametric body model for pose and shape and, a library of clothes ready to use in a single click, which allows us to accelerate the generation of a physics cloth dataset. From the cloth library we select three different 3D models (shirt, pants, dress) that will be used as our garment templates(tops, bottoms and dress). Each selected 3D model will be run with physical simulation activated for a total of $50$ actions taken from CMU \cite{CMU} and Mixamo \cite{Mixamo} mocap files.

Each simulation is designed bearing in mind that we want to capture the dynamics of a cloth during a long sequence action. In the current physics simulator based on spring-mass model \cite{Baraff1998}, the cloth behavior is influenced by different parameters that can be grouped in three main areas: 1) garment parameters $g_p$, 2) world parameters $w_p$ and 3) external forces parameters $f_p$. World parameters such as gravity and air friction are unchanged for all simulations. External forces such as velocity and acceleration parameters are constrained by the action defined in the motion files, and the garment parameters such as bending, stiffness, compression and shear, are adjusted to match a cotton fabric style simulation for each one of the cloth templates. The cloth fabric style needs to be adjusted for every cloth mesh used, as these parameters depends on the number of vertices of the mesh. The simulations are run with collisions and self-collisions activated.

\subsection{Generate train UV maps}

\begin{figure}[t]
\begin{center}
   \includegraphics[width=0.99\linewidth]{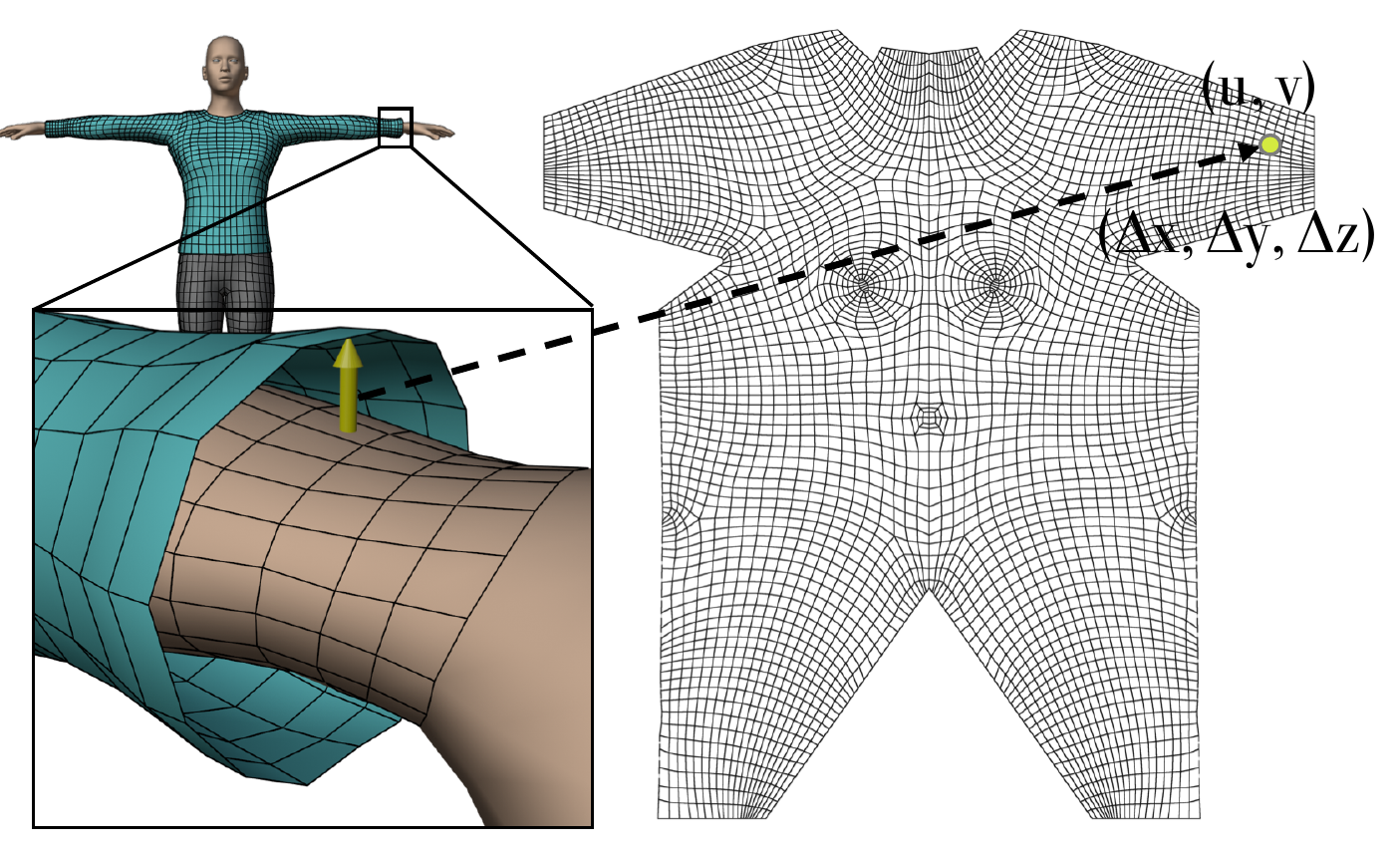}
\end{center}
\vspace{-0.5cm}
   \caption{Offset garment calculation. Given a point in the body surface with UV coordinates $u, v$, a ray is thrown in the direction of its normal until intersects the garment model. Then, the vector $\Delta x, \Delta y, \Delta z$ between the body and the garment is stored in the garment UV map.}
\label{fig:normals}
\end{figure}

\begin{figure}[t]
\begin{center}
   \includegraphics[width=0.99\linewidth]{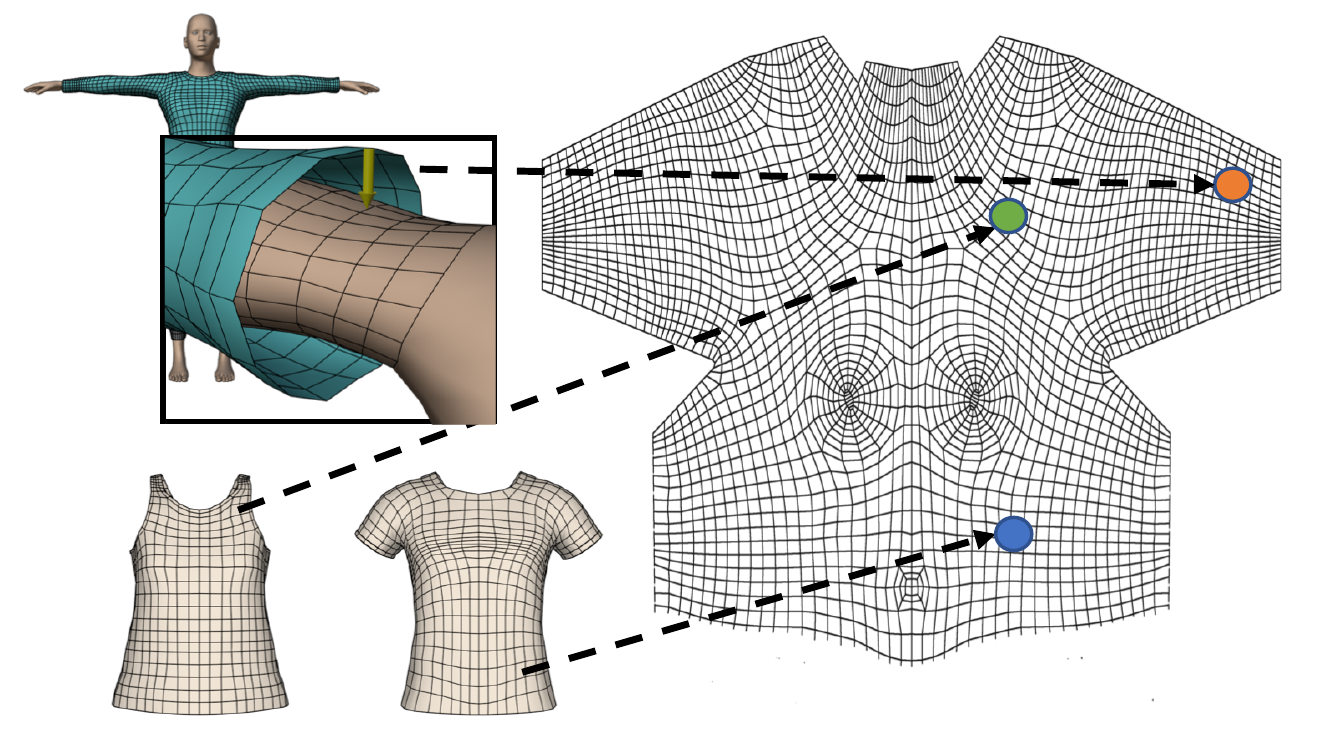}
\end{center}
\vspace{-0.5cm}
   \caption{Cloth mesh to garment template projection. An arbitrary garment is worn into the human body model in T-pose position. Then, for every vertex, a ray is thrown in the direction of its normal until intersects the body model. The point on the surface of the body model has a correspondence with the coordinates of the garment templates. }
\label{fig:projection}
\end{figure}

The synthetic dataset is generated from the 3D mesh models for body and clothes. The body mesh is a parametric model $\mathbf{M}^B_k (\alpha, \theta) \in \mathbb{R}^{3 \times N}$ with $N$ vertices and a set of parameters to control shape $\alpha$ and pose $\theta$ at frame sequence $k$. This body 3D model will wear each one of the three following cloth mesh templates, tops $\mathbf{M}^T_k \in \mathbb{R}^{3 \times M_t}$, bottoms $\mathbf{M}^{Bt}_k \in \mathbb{R}^{3 \times M_b}$ and dresses $\mathbf{M}^{Dr}_k \in \mathbb{R}^{3 \times M_d}$, with $M_t$, $M_b$, $M_d$ vertices respectively. For simplicity in the notation we will refer to the cloth mesh template models as $M^C_k$ when the models are at sequence frame $k$. As there is no direct correspondence between the vertices of the body mesh and the vertices of the cloth templates, we define a transference matrix $T^{BC}$ which relates the body vertices with a point in the cloth surface, see Fig. \ref{fig:normals} and $T^{CB}$ relates cloth vertices with a point in the body surface, see Fig. \ref{fig:projection}.

Hence, each cloth vertex position at frame $k$, can be expressed as a point in the body surface with an offset $\mathbf{O}_k$ as in Eq. \ref{eq:garment}:
\begin{align}
\mathbf{M}^C_k &= \mathbf{T}^{CB} \mathbf{M}^B_k(\alpha, \theta) + \mathbf{O}_k \label{eq:garment} \\
\mathbf{O}_k &= f_{P,k} (f_p, g_p, w_p) \label{eq:physics} 
\end{align}
where $f_{P,k}( b_f, g_p, w_p)$ is a function that defines the offset positions of each vertex given a set of parameters such as body forces $f_p$, world scene $w_p$ and garment fabric $g_p$.

\noindent{\bf Body UV maps.} Neural networks are more efficient when using 2D image representations, and for that reason we will represent our 3D models surface by means of a UV maps. Each pixel $(u, v)$ of the UV layout has a direct correspondence with a point in the mesh surface stored in the transference matrix $\mathbf{T}^{UB}$. Therefore, the body mesh surface is represented by the body UV map $\mathbf{I}^B_k (u, v)$. From the body UV map positions we can easily obtain the UV maps for velocity $\mathbf{I}^B_{v,k} (u, v)$ and acceleration $\mathbf{I}^B_{a,k} (u, v)$.
\begin{align}
\mathbf{I}^B_k (u, v) &= \mathbf{T}^{UB} (u, v) \mathbf{M}^B_k (\alpha, \theta)  \label{eq:body_uvmap} \\
\mathbf{I}^B_{v,k} (u, v)  &= \mathbf{I}^B_k - \mathbf{I}^B_{k-1}  \\
\mathbf{I}^B_{a,k} (u, v) &=  \mathbf{I}^B_{v,k} (u, v) - \mathbf{I}^B_{v,k-1} (u, v)
\end{align}

The original body UV map layout is modified to occupy as many pixels as possible inside the layout and therefore, get a better sampling of the surface of the body. Thus, face, hands and feet are removed from the layout, and upper and lower limbs are stretched and resized, see Fig. \ref{fig:normals}.

\noindent{\bf Garment UV maps.}  The garment UV maps $I^{C}_k (u, v)$ will contain the offset vectors from the body surface to the cloth surface points for each pixel in the transference matrix $\mathbf{T}^{BC} (u, v)$, Eq. \ref{eq:garment_uvmap}. This matrix is calculated with the body dressed with a T-pose position. Then, for every valid pixel in the body UV map $\mathbf{I}^B_k$, is traced a ray along the normal of the body surface and the impact point is stored as the cloth point correspondence. This process is illustrated in Fig. \ref{fig:normals}.
\begin{align}
\mathbf{I}^{C}_k (u, v) &= \mathbf{T}^{BC} (u, v) \mathbf{M}^{C}_k - \mathbf{M}^B_k (\alpha, \theta) \label{eq:garment_uvmap}
\end{align}

The case of the dress garment $\mathbf{I}^{Dr}_k (u, v)$ is a bit different, since in the lower part of the dress garment will be parts of the mesh that have no body correspondence due to the rays along the surface normals of the inside part of the leg never impact to the center of the garment. Therefore, another body mesh $M^{Bd}_k (\alpha, \theta)$ where the legs are joined by an ellipsoid  is created.

\begin{figure*}
\begin{center}
  \includegraphics[width=.9\linewidth]{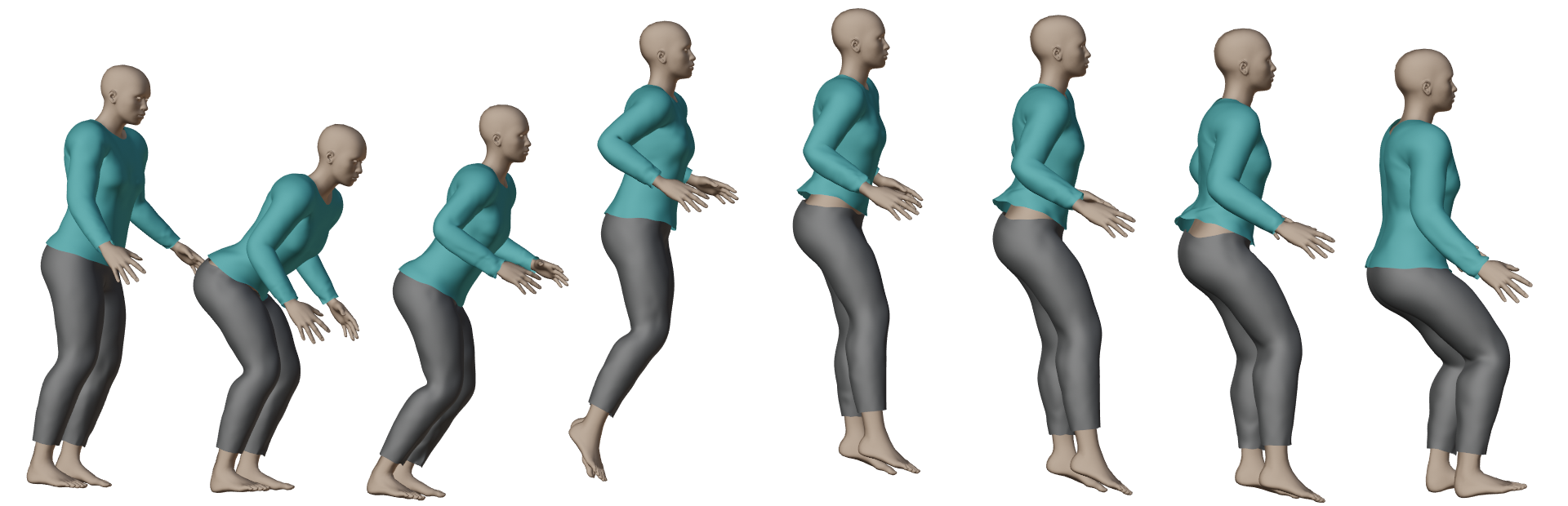} \\
  \includegraphics[width=.9\linewidth]{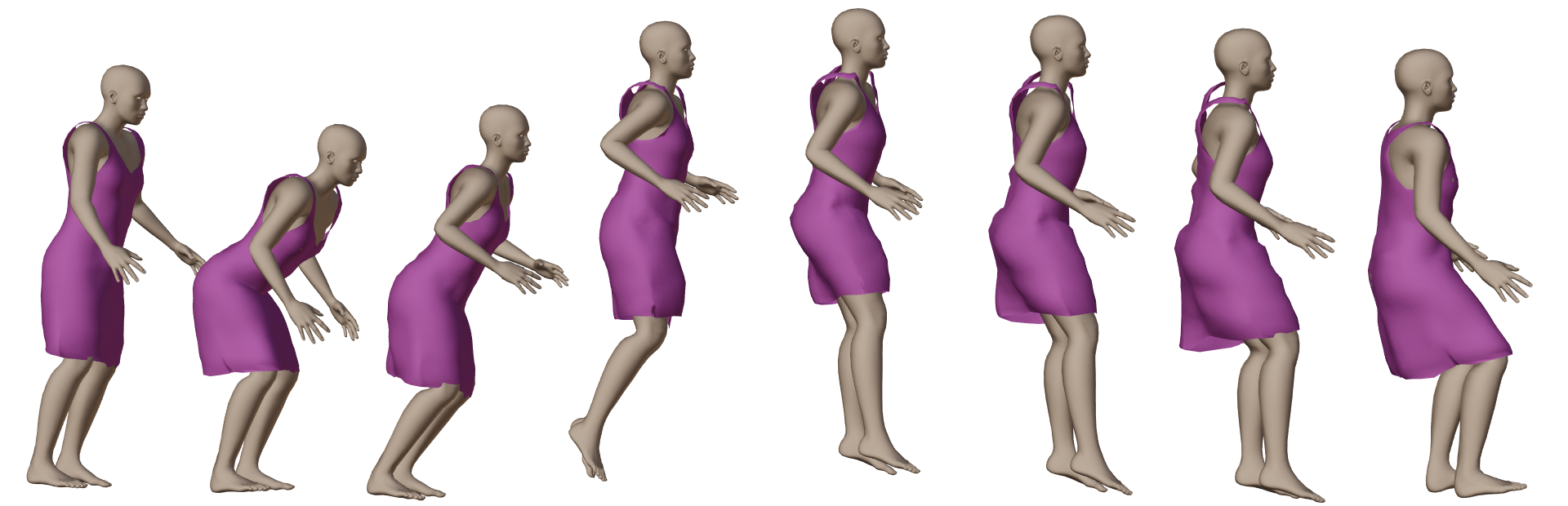} \\
\end{center}
\vspace{-0.5cm}
   \caption{\textbf{Qualitative results.} Several frames of the sequence action jump with tops and bottoms templates in the first row and dress template in the second row. 
   }
\label{fig:results_net}
\vspace{-3mm}
\end{figure*}

\subsection{Evaluate different garments}
The main advantage of the PhysXNet network over other methods is that we can easily use garments from different sources without the need of retrain the network. These garments need to be able to be encapsulated in one of the three cloth templates, but there is no condition about the number of vertices neither the topology.

Thus, given a garment model $\mathbf{M}^X_k \in \mathbb{R}^{3 \times N_x}$ where $N_x$ is an arbitrary number of vertices we need to find the transference matrix $\mathbf{T}^{XB}$ that relates a vertex of the model with the garment templates $\mathbf{I}^C$. This process illustrated in Fig. \ref{fig:projection}, and consists into throw a ray from the cloth vertex $N_x$ along its normal direction to the body surface in a T-pose, in order to find the body UV map coordinate $(u, v)$. The body UV map $\mathbf{I}^B_k$ has a direct correspondence with each one of the estimated garment templates by the transference matrix $\mathbf{T}^{BC}$ computed previously. This coordinate will give us the offset respect to the body, and we will be able to reconstruct the mesh for a given frame $k$.
\begin{align}
\textbf{M}^{X}_k  &= \mathbf{T}^{XB} \mathbf{M}^{B}_k (\alpha, \theta) + \mathbf{O}_k 
\end{align}

In our case, the vertices of the cloth meshes evaluated fall a few pixels far from the UV map boundaries. This fact, avoid us some discrepancies that could be in the opposite UV map coordinates values corresponding to the same vertex. In case a vertex fall in the UV map boundary, the best solution would be to average pixel values corresponding the same vertex.

\subsection{Implementation details}

The PhysXNet network is trained with $22$ actions with a total of $3,217$ frames and tested with $12$ actions with a total of $1,620$ frames. All data UV maps, $\mathbf{I}^B_{v,k}$, $\mathbf{I}^B_{a,k}$, $\mathbf{I}^C_k$ are normalized independently from $-1$ to $1$. The network discriminator is trained with soft labels, using random uniform sampling from $0.0$ to $0.3$ for estimated labels, and from $0.7$ to $1.0$ for ground truth labels. Moreover, a random $5\%$ of training data on each epoch contain flip labels. Image UV map sizes are $W = H = 256$, $\lambda_{L_1} = 100$ and learning rate $r = 2e-04$.
The architecture is trained up to $150$ epochs for $2$ days in a single NVIDIA GeForce GTX 1080 GPU and inference mean time per frame is $0.0313s$ (load data, run, save files).

\section{Experiments}

We next evaluate our proposed PhysXNet performing several quantitative and qualitative experiments. In the quantitative experiments, we compare our proposed method with the Linear Blend Skinning (LBS) method as a baseline. The LBS method calculates the displacement of each vertex according to a weighted linear combination of the assigned skeleton segments. Results are given by comparing the estimated UV garment maps with the ground truth UV maps for each vertex of the garment template and also for each pixel in the UV garment map. 
In the qualitative results, we compare our proposed method with LBS and TailorNet \cite{Patel_2020_CVPR}. We also show the results of PhysXNet with different body shapes and other garment meshes than the ones used for train.

\subsection{Quantitative results}

We provide two different measures for the quantitative results. First, we calculate the mean squared error (MSE) for each valid pixel of the PhysXNet estimated UV map templates $\hat{\mathbf{I}}^C_{0,k}$ with the ground truth UV maps obtained from the synthetic dataset $\mathbf{I}^C_{0,k}$. Then, we calculate the MSE error for the vertices of the cloth meshes used to generate the dataset. Note that the meshes are a subset of the UV map pixels.

\begin{figure}[t]
\begin{center}
  \subfloat[MSE (mm.) UV map templates]{\includegraphics[width=\linewidth]{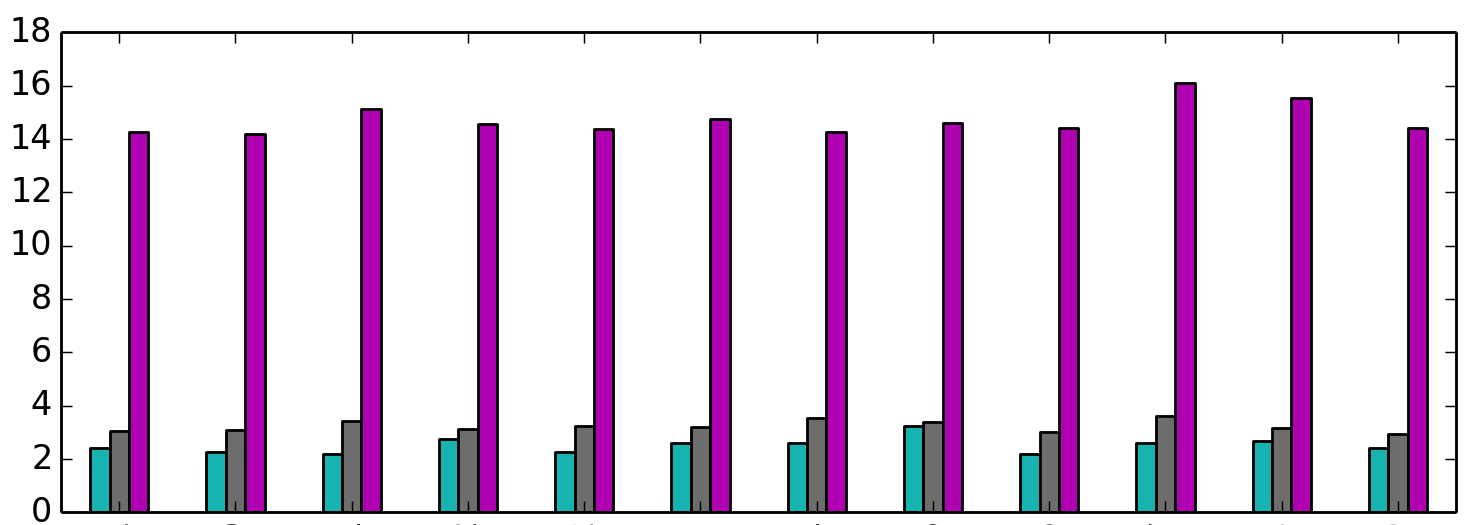} } \\
  \subfloat[MSE (mm.) vertices mesh templates]{\includegraphics[width=\linewidth]{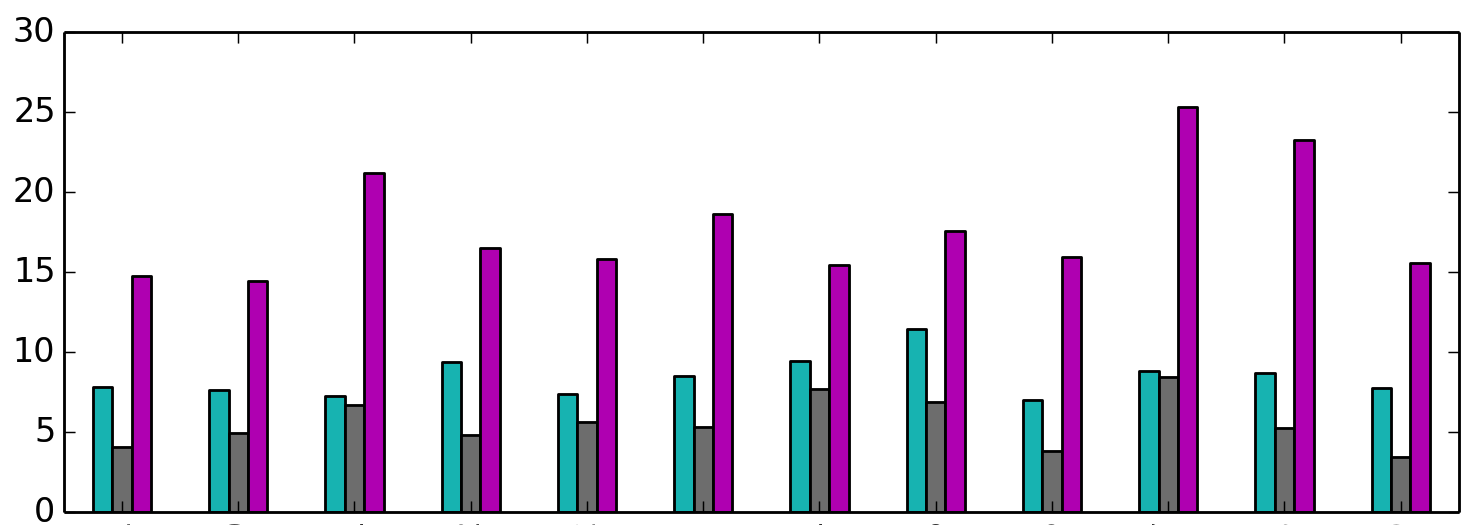} } 

\end{center}
\vspace{-0.5cm}
   \caption{\textbf{Quantitative results.} Mean squared error (MSE) in mm. for UV map estimated templates (above) and for mesh template vertices (below). 
   }
\label{fig:quantitative_results}
\vspace{-3mm}
\end{figure}

The evaluated actions in the Fig.~\ref{fig:quantitative_results} are in the following order: \em{ jump}, \em{walking}, \em{moon walk}, \em{Chinese dance}, \em{punch}, \em{balancing}, \em{ballet}, \em{stretch arms}, \em{salsa dance}, \em{jogging}, \em{side step}, and \em{strong gesture}. The first bar with color cyan is for the tops template, the second bar with gray color is for the bottoms template and the third bar with purple color is for the dress template. Some of these actions have very soft movements, like \em{moon walk}, \em{balancing}, \em{walking}, which results in small velocities and accelerations while some of the other motions like \em{strong gesture}, \em{jump}, \em{punch} in very few frames the pose has big changes which produces large velocities and accelerations.

Errors for tops and bottoms templates are very similar while errors in the dress template are almost double than the other two. The reason why the dress template errors are bigger is due to the hallucination that the network needs to do in the legs of the body, as there are parts of the dress that have no direct correspondence with the input body UV maps.

\begin{figure} [!ht]
\begin{center}
  \includegraphics[width=\linewidth]{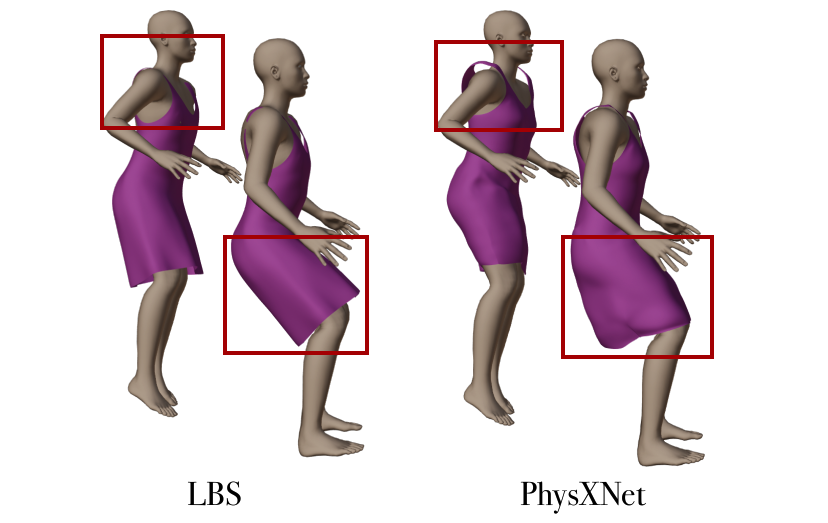} 
\end{center}
\vspace{-0.5cm}
   \caption{\textbf{Qualitative results.} Comparison for LBS and PhysXNet methods. Here we find main differences in the upper part and bottom part of the dress. LBS method the dress just follows the body movements while our proposed method transfers the inertia of the movement to the dress. Note that TailorNet has no dress garment, therefore there is no possible comparison.
   }
\label{fig:results_comparison2}
\vspace{-3mm}
\end{figure}

\subsection{Qualitative results}

In the qualitative results we compare our proposed method PhysXNet, with LBS method and TailorNet \cite{Patel_2020_CVPR} when possible. Moreover, we show the performance under changes of human shape and different cloth topologies that have never been seen by the network.

\textbf{Results on continuous actions.} We show several frames for the sequence \textit{Jump} in Fig. \ref{fig:results_net}. In the top row the model wears the tops and bottoms templates, and in the bottom row, the model wears the dress template. We can observe that when the body goes up it generates several forces that push the cloth also up. This kind of behaviour is not possible to obtain with other methods that only use human pose to deform the cloth.

\begin{figure*} [!ht]
\begin{center}
  \includegraphics[width=\linewidth]{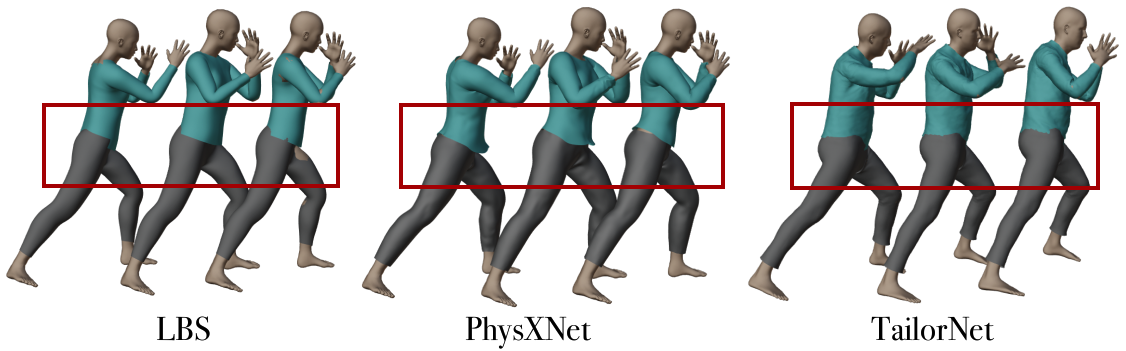} \\
\end{center}
\vspace{-0.5cm}
   \caption{\textbf{Qualitative results.} Comparison for the LBS, PhysXNet and TailorNet methods for the action Punch. Most differences can be found in the bottom part of the shirt. The proposed PhysXNet is able to model the movement of the bottom part of the shirt during the action, while other two methods keep it on the same position for all the frames. Note that TailorNet results are calculated on a different 3D body model. 
   }
\label{fig:results_comparison}
\vspace{-3mm}
\end{figure*}

\begin{figure} [t!]
\begin{center}
  \includegraphics[width=\linewidth]{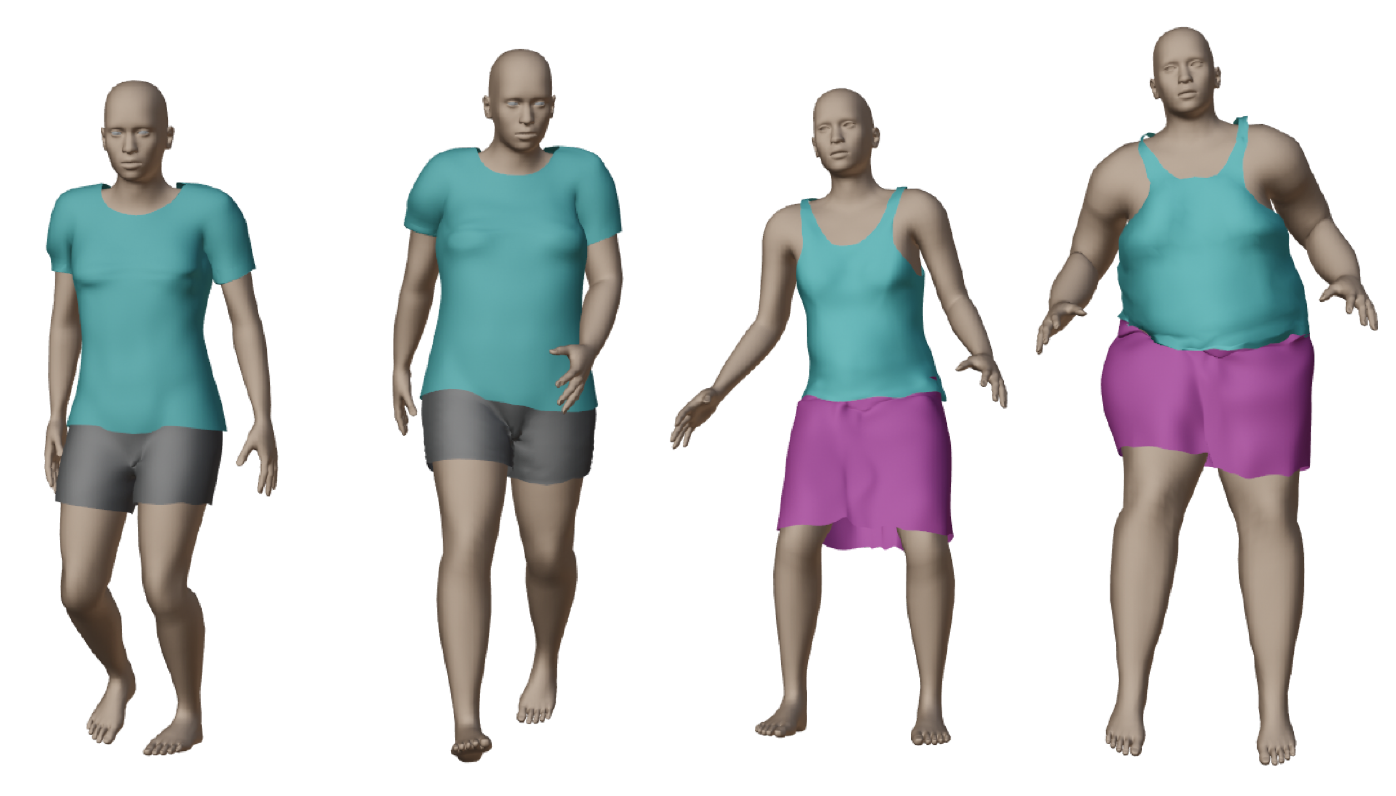} \\
\end{center}
\vspace{-0.5cm}
   \caption{\textbf{Qualitative results.} Examples of different body shapes using several cloth meshes that have never been seen by the network.
   }
\label{fig:results_shape}
\vspace{-0.5cm}
\end{figure}

\textbf{Different garment topologies and body shapes.} The proposed PhysXNet is also able to deal with different body shapes, as the output of the network are the offsets of each garment respect to the body, and also with cloth garments that contains different number of vertices and different topologies. This is possible due to the output UV map templates encode the surface of the garment, and when using a different mesh topology, it is only necessary to project the vertices of the mesh with the UV map without being necessary to retrain the network. Results for 3D cloth (t-shirt, shorts, shirt2, skirt) are shown in Fig. \ref{fig:results_shape}.

\textbf{Comparisons.} We compare our PhysXNet network with the LBS method and TailorNet \cite{Patel_2020_CVPR} in the case of Tops and Bottoms templates, and only with LBS method in the case of Dress template, as TailorNet does not have a dress model. The human models used in our case, Makehuman \cite{Makehuman}, and TailorNet, SMPL \cite{SMPL_2015}, are different, and this makes that the represented actions are not exactly the same due to the internal bone structures and bone lengths.

In Fig. \ref{fig:results_comparison} we can observe the differences between the three methods. While in the LBS and TailorNet methods, the bottom part of the shirt is not moving while the performing \textit{punch} action, in our proposed method, the shirt contains the movement produced by the body movements. The main reason for this behavior is because our method takes into account the current and past body motion and is able to apply it to the cloth, while other two methods are static and only use current body pose. Similar we can observe in Fig. \ref{fig:results_comparison2} with the dress template.

There are also other differences between the three methods shown in Table \ref{table:comparisons}. First, about the garment mesh models itself. While TailorNet uses very heavy models, up to 14GB per model, our cloth models and LBS are very light, being around 1MB. The counterpart, is that our model is very widely sampled, making difficult to capture small wrinkles. A second difference is about the network weights, while for a single garment, network weights are similar in size, in TailorNet, if a user wants to use more models it is necessary to download more weights. In our proposed method, the same weights are used for the three templates which can be applied to a large variety of garments. The last difference is about the execution time, as expected larger models comes with larger execution times. Hence, in PhysXNet, with a single pass of the network we obtain the outputs of the three garment templates, in TailorNet it is be necessary to perform inference for each one of the desired garments.

\begin{table}[t!]
\centering
\begin{tabular}{c c c c c}
     & Cloth & LBS & TailorNet & \textbf{PhysXNet}  \\
    \toprule
    3D Model & Shirt & 1MB & 13.5GB  & 1MB  \\
     & Pants & 1MB & 3.4GB  & 1MB \\
    \midrule
    Network & Shirt & -- & 2.5GB  & 1GB \\
    Weights & Pants & -- & 1.2GB  & 1GB \\
    \midrule 
    Inference & Shirt & 3ms & 1s  & 30ms \\
    Time & Pants & 3ms & 500ms  & 30ms \\
    \bottomrule
\end{tabular}
\caption{Comparison of 3D model size, network weights size and inference time for the LBS, TailorNet and PhysXNet methods. Note that the proposed PhysXNet method can deal with multiple cloth templates at the same time, hence, network weights size, and inference time are much lower than TailorNet method.}
\label{table:comparisons}
\vspace{-0.5cm}
\end{table}

\section{Conclusions}
\vspace{-1mm}
We presented a network, PhysXNet, that generates cloth physical dynamics for three totally different garment templates at the same time. The network is able to generalize to unseen body actions, different body shapes and different cloth 3D models, making the model suitable to integrate it into a larger pipeline. Our network can simulate the cloth physics behavior for any 3D cloth mesh randomly downloaded from the internet that fits to any of the three garment templates without being retrained. 
The proposed method is compared quantitatively with the synthetic dataset ground truth and qualitatively with a baseline, LBS, and to an state-of-the-art method, TailorNet.

\section*{Acknowledgment}
\vspace{-1mm}
This work is supported partly by the Spanish government under  project MoHuCo PID2020-120049RB-I00, the ERA-Net Chistera project IPALM PCI2019-103386 and María de Maeztu Seal of Excellence MDM-2016-0656.

{\small
\bibliographystyle{ieee_fullname}
\bibliography{refs}
}

\end{document}

%% file: newcommands2.tex
\newcommand{\comment}[1]{}

\newcommand{\bI}{\mathbf{I}}

\newcommand{\bX}{\mathbf{X}}
\newcommand{\bY}{\mathbf{Y}}

\newcommand{\mM}{\mathcal{M}}




%% file: physxnet_camready.bbl
\begin{thebibliography}{10}\itemsep=-1pt

\bibitem{Baraff1998}
David Baraff and Andrew Witkin.
\newblock Large steps in cloth simulation.
\newblock In {\em SIGGRAPH '98: Proceedings of the 25th annual conference on
  Computer graphics and interactive techniques}, pages 43--54, 1998.

\bibitem{bertiche2020cloth3d}
Hugo Bertiche, Meysam Madadi, and Sergio Escalera.
\newblock Cloth3d: Clothed 3d humans, 2020.

\bibitem{bhatnagar2019mgn}
Bharat~Lal Bhatnagar, Garvita Tiwari, Christian Theobalt, and Gerard Pons-Moll.
\newblock Multi-garment net: Learning to dress 3d people from images.
\newblock In {\em {IEEE} International Conference on Computer Vision ({ICCV})}.
  {IEEE}, oct 2019.

\bibitem{Blender}
Blender.
\newblock \url{https://www.blender.org}.
\newblock Accessed: 2020-09-30.

\bibitem{CMU}
Cmu.
\newblock \url{http://mocap.cs.cmu.edu}.
\newblock Accessed: 2020-09-30.

\bibitem{corona2021smplicit}
Enric Corona, Albert Pumarola, Guillem Aleny{\`a}, Gerard Pons-Moll, and
  Francesc Moreno-Noguer.
\newblock Smplicit: Topology-aware generative model for clothed people.
\newblock In {\em Proceedings of the IEEE/CVF Conference on Computer Vision and
  Pattern Recognition (CVPR)}, 2021.

\bibitem{Geng2020tshirt}
Zhenglin Geng, Daniel Johnson, and Ronald Fedkiw.
\newblock Coercing machine learning to output physically accurate results.
\newblock {\em Journal of Computational Physics}, 406, 2020.

\bibitem{Goodfellow-NIPS-2014}
Ian~J. Goodfellow, Jean Pouget-Abadie, Mehdi Mirza, Bing Xu, David
  Warde-Farley, Sherjil Ozair, Aaron Courville, and Yoshua Bengio.
\newblock Generative adversarial nets.
\newblock In {\em NIPS}, 2014.

\bibitem{Guan-DRAPE-2012}
Peng Guan, Loretta Reiss, David~A. Hirshberg, Alexander Weiss, and Michael~J.
  Black.
\newblock Drape: Dressing any person.
\newblock {\em ACM Trans. Graph.}, 31(4), 2012.

\bibitem{Gundogdu-ICCV-2019}
Erhan Gundogdu, Victor Constantin, Amrollah Seifoddini, Minh Dang, Mathieu
  Salzmann, and Pascal Fua.
\newblock Garnet: A two-stream network for fast and accurate 3d cloth draping.
\newblock In {\em {IEEE} International Conference on Computer Vision ({ICCV})},
  2019.

\bibitem{habermann20deepcap}
Marc Habermann, Weipeng Xu, , Michael Zollhoefer, Gerard Pons-Moll, and
  Christian Theobalt.
\newblock Deepcap: Monocular human performance capture using weak supervision.
\newblock In {\em {IEEE} Conference on Computer Vision and Pattern Recognition
  (CVPR)}. {IEEE}, jun 2020.

\bibitem{pix2pix2017}
Phillip Isola, Jun-Yan Zhu, Tinghui Zhou, and Alexei~A Efros.
\newblock Image-to-image translation with conditional adversarial networks.
\newblock In {\em CVPR}, 2017.

\bibitem{jin2018pixelbased}
Ning Jin, Yilin Zhu, Zhenglin Geng, and Ronald Fedkiw.
\newblock A pixel-based framework for data-driven clothing, 2018.

\bibitem{Kato2019clothes}
Natsumi Kato, Hiroyuki Osone, Kotaro Oomori, Chun Ooi, and Yoichi Ochiai.
\newblock Gans-based clothes design: Pattern maker is all you need to design
  clothing.
\newblock In {\em Augmented Human International Conference (AH)}, 2019.

\bibitem{Koh_2015_VDA}
Woojong Koh, Rahul Narain, and James~F. O'Brien.
\newblock View-dependent adaptive cloth simulation with buckling compensation.
\newblock {\em IEEE Transactions on Visualization and Computer Graphics},
  21(10), 2015.

\bibitem{li2020deep}
Yue Li, Marc Habermann, Bernhard Thomaszewski, Stelian Coros, Thabo Beeler, and
  Christian Theobalt.
\newblock Deep physics-aware inference of cloth deformation for monocular human
  performance capture, 2020.

\bibitem{SMPL_2015}
Matthew Loper, Naureen Mahmood, Javier Romero, Gerard Pons-Moll, and Michael~J.
  Black.
\newblock {SMPL}: A skinned multi-person linear model.
\newblock {\em ACM Trans. Graphics (Proc. SIGGRAPH Asia)}, 34(6), 2015.

\bibitem{laehner2018deepwrinkles}
Z. Lähner, D. Cremers, and T. Tung.
\newblock Deepwrinkles: Accurate and realistic clothing modeling.
\newblock In {\em European Conference on Computer Vision (ECCV)}, 2018.

\bibitem{ma20autoenclother}
Qianli Ma, Jinlong Yang, Anurag Ranjan, Sergi Pujades, Gerard Pons-Moll, Siyu
  Tang, and Michael Black.
\newblock Learning to dress 3d people in generative clothing.
\newblock In {\em {IEEE} Conference on Computer Vision and Pattern Recognition
  (CVPR)}. {IEEE}, jun 2020.

\bibitem{Makehuman}
Makehuman.
\newblock \url{http://www.makehumancommunity.org}.
\newblock Accessed: 2020-09-30.

\bibitem{Mixamo}
Mixamo software.
\newblock \url{https://www.mixamo.com}.
\newblock Accessed: 2020-09-30.

\bibitem{Moreno_pami2013}
F. Moreno-Noguer and P. Fua.
\newblock Stochastic exploration of ambiguities for nonrigid shape recovery.
\newblock {\em IEEE Transactions on Pattern Analysis and Machine Intelligence
  (PAMI)}, 35(2):463--475, 2013.

\bibitem{Moreno_cvpr2009}
F. Moreno-Noguer, M. Salzmann, V. Lepetit, and P. Fua.
\newblock Capturing 3d stretchable surfaces from single images in closed form.
\newblock In {\em Proceedings of the Conference on Computer Vision and Pattern
  Recognition (CVPR)}, pages 1842--1849, 2009.

\bibitem{Patel_2020_CVPR}
Chaitanya Patel, Zhouyingcheng Liao, and Gerard Pons-Moll.
\newblock Tailornet: Predicting clothing in 3d as a function of human pose,
  shape and garment style.
\newblock In {\em Proceedings of the IEEE/CVF Conference on Computer Vision and
  Pattern Recognition (CVPR)}, June 2020.

\bibitem{pumarola2020d}
Albert Pumarola, Enric Corona, Gerard Pons-Moll, and Francesc Moreno-Noguer.
\newblock D-nerf: Neural radiance fields for dynamic scenes.
\newblock In {\em Proceedings of the IEEE/CVF Conference on Computer Vision and
  Pattern Recognition (CVPR)}, 2021.

\bibitem{pumarola20193dpeople}
Albert Pumarola, Jordi Sanchez-Riera, Gary Choi, Alberto Sanfeliu, and Francesc
  Moreno-Noguer.
\newblock {3DPeople: Modeling the Geometry of Dressed Humans}.
\newblock In {\em International Conference in Computer Vision (ICCV)}, 2019.

\bibitem{Ronnegerger2015unet}
O. Ronneberger, P.Fischer, and T. Brox.
\newblock U-net: Convolutional networks for biomedical image segmentation.
\newblock In {\em Medical Image Computing and Computer-Assisted Intervention
  (MICCAI)}, volume 9351, 2015.

\bibitem{Sanchez_cvpr2010}
J. Sanchez, J. Östlund, P. Fua, and F. Moreno-Noguer.
\newblock Simultaneous pose, correspondence and non-rigid shape.
\newblock In {\em Proceedings of the Conference on Computer Vision and Pattern
  Recognition (CVPR)}, pages 1189--1196, 2010.

\bibitem{sanchezriera2021avatar}
Jordi Sanchez-Riera, Aniol Civit, Marta Altarriba, and Francesc Moreno-Noguer.
\newblock Avatar: Blender add-on for fast creation of 3d human models, 2021.

\bibitem{Santesteban-EG-2019}
Igor Santesteban, Miguel~A. Otaduy, and Dan Casas.
\newblock {Learning-Based Animation of Clothing for Virtual Try-On}.
\newblock {\em Computer Graphics Forum}, 2019.

\bibitem{santesteban2021garmentcollisions}
Igor Santesteban, Nils Thuerey, Miguel~A Otaduy, and Dan Casas.
\newblock {Self-Supervised Collision Handling via Generative 3D Garment Models
  for Virtual Try-On}.
\newblock {\em IEEE/CVF Conference on Computer Vision and Pattern Recognition
  (CVPR)}, 2021.

\bibitem{shen2020garmentgeneration}
Yu Shen, Junbang Liang, and Ming~C. Lin.
\newblock Gan-based garment generation using sewing pattern images.
\newblock In {\em Proceedings of the European Conference on Computer Vision
  (ECCV)}, 2020.

\bibitem{Tang2016Cama}
Min Tang, Huamin Wang, Le Tang, Ruofeng Tong, and Dinesh Manocha.
\newblock Cama: Contact-aware matrix assembly with unified collision handling
  for gpu-based cloth simulation.
\newblock {\em Computer Graphics Forum}, 35:511--521, 05 2016.

\bibitem{Tang2018TOG}
Min Tang, TongTong Wang, Zhongyuan Liu, Ruofeng Tong, and Dinesh Manocha.
\newblock I-cloth: Incremental collision handling for gpu-based interactive
  cloth simulation.
\newblock {\em ACM Transactions on Graphics}, 37(6), 2018.

\bibitem{SIZER_Dataset}
Garvita Tiwari, Bharat Bhatnagar, Tony Tung, and Gerard Pons-Moll.
\newblock Sizer: A dataset and model for parsing 3d clothing and learning size
  sensitive 3d clothing.
\newblock In {\em European Conference on Computer Vision (ECCV)}, 2020.

\bibitem{varol17_surreal}
G{\"u}l Varol, Javier Romero, Xavier Martin, Naureen Mahmood, Michael~J. Black,
  Ivan Laptev, and Cordelia Schmid.
\newblock Learning from synthetic humans.
\newblock In {\em CVPR}, 2017.

\bibitem{vidaurre2020virtualtryon}
Raquel Vidaurre, Igor Santesteban, Elena Garces, and Dan Casas.
\newblock {Fully Convolutional Graph Neural Networks for Parametric Virtual
  Try-On}.
\newblock {\em Computer Graphics Forum (Proc. SCA)}, 2020.

\bibitem{garmentdesign_Wang_SA18}
Tuanfeng~Y. Wang, Duygu Ceylan, Jovan Popovic, and Niloy~J. Mitra.
\newblock Learning a shared shape space for multimodal garment design.
\newblock {\em ACM Trans. Graph.}, 37(6), 2018.

\bibitem{WangEtAl-SiggraphAsia-2019}
Yangtuanfeng Wang, Tianjia Shao, Kai Fu, and Niloy Mitra.
\newblock Learning an intrinsic garment space for interactive authoring of
  garment animation.
\newblock {\em ACM Trans. Graph.}, 38(6), 2019.

\bibitem{Yang2018garments}
Shan Yang, Zherong Pan, Tanya Amert, Ke Wang, Licheng Yu, Tamara Berg, and
  Ming~C. Lin.
\newblock Physics-inspired garment recovery from a single-view image.
\newblock {\em ACM Transactions on Graphics}, 37(5), 2018.

\bibitem{SimulCap19}
Tao Yu, Zerong Zheng, Yuan Zhong, Jianhui Zhao, Dai Quionhai, Gerard Pons-Moll,
  and Yebin Liu.
\newblock Simulcap : Single-view human performance capture with cloth
  simulation.
\newblock In {\em {IEEE} Conference on Computer Vision and Pattern Recognition
  (CVPR)}, jun 2019.

\bibitem{zhang2020deep}
Meng Zhang, Tuanfeng Wang, Duygu Ceylan, and Niloy~J Mitra.
\newblock Deep detail enhancement for any garment.
\newblock {\em arXiv preprint arXiv:2008.04367}, 2020.

\end{thebibliography}
